\documentclass[11pt,a4paper]{article}
\usepackage[hyperref]{emnlp2018}
\usepackage{times}
\usepackage{latexsym}
\usepackage{subcaption}
\usepackage{url}
\usepackage{blindtext}
\usepackage{graphicx}
\usepackage{comment}
\graphicspath{ {images/} }
\usepackage{booktabs}

\usepackage{caption}
\usepackage{dblfloatfix}

\aclfinalcopy % Uncomment this line for the final submission
\title{A Word-Complexity Lexicon and A Neural Readability Ranking Model for Lexical Simplification}

\author{Mounica Maddela \and Wei Xu \\ Department of Computer Science and Engineering \\
The Ohio State University\\
  {\tt \{maddela.4, xu.1265\}@osu.edu}
}

\date{}

\begin{document}
\maketitle
\begin{abstract}
Current lexical simplification approaches rely heavily on heuristics and corpus level features that do not always align with human judgment. We create a human-rated word-complexity lexicon of 15,000 English words and propose a novel neural readability ranking model with a Gaussian-based feature vectorization layer that utilizes these human ratings to measure the complexity of any given word or phrase. Our model performs better than the state-of-the-art systems for different lexical simplification tasks and evaluation datasets. Additionally, we also produce SimplePPDB++, a lexical  resource of over 10 million simplifying paraphrase rules, by applying our model to the Paraphrase Database (PPDB).\footnote{The code and data are publicly available on the authors' homepages and GitHub: \url{https://github.com/mounicam/lexical_simplification}.}
\end{abstract}

\section{Introduction}

Lexical simplification is an important subfield that is concerned with the complexity of words or phrases, and particularly how to measure readability and reduce the complexity using alternative paraphrases. There are three major lexical simplification tasks which effectively resemble a pipeline: (i) Complex Word Identification \cite{paetzold2016a,Muhie2017, Shardlow2013a} which involves identifying complex words in the sentence; (ii) Substitution Generation \cite{glavavs-vstajner:2015:ACL-IJCNLP, Coster:2011:LSS:2107679.2107680} which involves finding alternatives to complex words or phrases; and (iii) Substitution Ranking \cite{specia2012} which involves ranking the paraphrases by simplicity. Lexical simplification also has practical real-world uses, such as displaying alternative expressions of complex words as reading assistance for children \cite{Kajiwara2013}, non-native speakers \cite{petersen2007text,pellow-eskenazi:2014:PITR}, lay readers \cite{Elhadad2007,siddharthan-katsos:2010:NAACLHLT}, or people with reading disabilities \cite{Rello:2013:ILS:2458308.2458354}.

Most current approaches to lexical simplification heavily rely on corpus statistics and surface level features, such as word length and corpus-based word frequencies (read more in \S \ref{sec:relatedwork}). Two of the most commonly used assumptions are that simple words are associated with shorter lengths and higher frequencies in a corpus. However, these assumptions are not always accurate and are often the major source of errors in the simplification pipeline \cite{shardlow2014}. For instance, the word \textit{foolishness} is simpler than its meaning-preserving substitution \textit{folly} even though \textit{foolishness} is longer and less frequent in the Google 1T Ngram corpus \cite{brants2006a}. In fact, we found that 21\% of the 2272 meaning-equivalent word pairs randomly sampled from PPDB\footnote{PPDB is a large paraphrase database derived from static bilingual translation data available at: \url{http://paraphrase.org}} \cite{ganitkevitch-EtAl:2013:NAACL} had the simpler word longer than the complex word, while 14\% had the simpler word less frequent.

%\footnote{PPDB is a large paraphrase database derived from static bilingual translation data. SimplePPDB is a subset of PPDB that contain around 4.5 million complex-to-simple English paraphrases automatically selected. Both are available at: \url{http://paraphrase.org}}

To alleviate these inevitable shortcomings of corpus and surface-based methods, we explore a simple but surprisingly unexplored idea -- creating an English lexicon of 15,000 words with word-complexity ratings by humans. We also propose a new neural readability ranking model with a Gaussian-based feature vectorization layer, which can effectively exploit these human ratings as well as other numerical features to measure the complexity of any given word or phrase (including those outside the lexicon and/or with sentential context). Our model significantly outperforms the state-of-the-art on the benchmark SemEval-2012 evaluation for Substitution Ranking \cite{specia2012,paetzold2017a}, with or without using the manually created word-complexity lexicon, achieving a Pearson correlation of 0.714 and 0.702 respectively. We also apply the new ranking model to identify lexical simplifications (e.g., \textit{commemorate} $\rightarrow$ \textit{celebrate}) among the large number of paraphrase rules in PPDB with improved accuracy compared to previous work for Substitution Generation. At last, by utilizing the word-complexity lexicon, we establish a new state-of-the-art on two common test sets for Complex Word Identification \cite{paetzold2016a,Muhie2017}. We make our code, the word-complexity lexicon, and a lexical resource of over 10 million paraphrase rules with improved readability scores (namely SimplePPDB++) all publicly available.

\section{Constructing A Word-Complexity Lexicon with Human Judgments}

We first constructed a lexicon of 15,000 English words with word-complexity scores assessed by human annotators.\footnote{Download at \url{https://github.com/mounicam/lexical_simplification}} Despite the actual larger English vocabulary size, we found that rating the most frequent 15,000 English words in Google 1T Ngram Corpus\footnote{\url{https://catalog.ldc.upenn.edu/ldc2006t13}} was effective for simplification purposes (see experiments in \S \ref{sec:experiments}) as our neural ranking model (\S \ref{sec:model}) can estimate the complexity of any word or phrase even out-of-vocabulary.

%\footnote{A recent online survey suggested that non-native English speakers most commonly have a vocabulary size of around 5,000 words and tend to reach over 10,000 if living aboard \cite{website:testyourvocab}. See also the interesting discussion on the difficulty of estimating how many words people know by Kenneth Churchz \shortcite{Church:2013:MME:2483691.2483693}.}

We asked 11 non-native but fluent English speakers to rate words on a 6-point Likert scale. We found that an even number 6-point scale worked better than a 5-point scale in a pilot experiment with two annotators, as the 6-point scheme allowed annotators to take a natural two-step approach: first determine whether a word is simple or complex; then decide whether it is `very simple' (or `very complex'), `simple' (or `complex'), or `moderately simple' (or `moderately complex'). For words with multiple capitalized versions (e.g., \textit{nature}, \textit{Nature}, \textit{NATURE}), we displayed the most frequent form to the annotators. We also asked the annotators to indicate the words for which they had trouble assessing their complexity due to ambiguity, lack of context or any other reason. All the annotators reported little difficulty, and explained possible reasons such as that word \textit{bug} is simple regardless of its meaning as an \textit{insect} in biology or an \textit{error} in computer software.\footnote{\label{note1}The word-happiness lexicon \cite{10.1371/journal.pone.0026752} of 10,222 words was also similarly created by human rating on the most frequent words without context or word-sense disambiguation.}

%We did not lemmatize the words because the inflectional forms may differ in complexity level from their lemmas. While partitioning the data, we also made sure that the words with the same lemma are in the same partition bu appear at random position.

\begin{table}[t!]
\centering
\small
\begin{tabular}{l|c|cccccc}
\hline\\[-1.0em]
\textbf{Word} & \textbf{Avg} & \textbf{A1} & \textbf{A2} & \textbf{A3} & \textbf{A4} & \textbf{A5} \\
\hline\\[-1.0em]
\textit{watch} & 1.0 & 1 & 1 & 1 & 1 & 1\\
\textit{muscles} & 1.6 & 2 & 1 & 2 & 2 & 1\\
\textit{sweatshirts} & 1.8 & 2 & 1 & 2 & 3 & 1\\
\textit{giant} & 2.0 & 2 & 3 & 1 & 1 & 3 \\
\textit{pattern} & 2.4 & 2 & 3 & 2 & 3 & 2\\
\textit{Christianity} & 2.8 & 3 & 2 & 2 & 3 & 4\\
\textit{educational} & 3.2 & 3 & 3 & 3 & 3 & 4\\
\textit{revenue} & 3.6 & 4 & 4 & 3 & 3 & 4\\
\textit{cortex} & 4.2 & 4 & 4 & 4 & 4 & 5\\
\textit{crescent} & 4.6 & 5 & 5 & 5 & 5 & 3 \\
%\textit{Obituaries} & 5.2 & 4 & 5 & 6 & 6 & 5\\
\textit{Memorabilia} & 5.4 & 5 & 6 & 6 & 5 & 5\\
\textit{assay} & 5.8 & 6 & 6 & 6 & 5 & 6\\

\hline
\end{tabular}
\small
\caption{Word-Complexity lexicon consists of English words and their complexity scores obtained by averaging over human ratings. A1, A2, A3, A4 and A5 are ratings by five different annotators on a 6-point Likert scale (1 is the simplest and 6 is the most complex).}
\label{table:lex}
\end{table}

With our hired annotators, we were able to have most annotators complete half or the full list of 15,000 words for better consistency, and collected between 5 and 7 ratings for each word. It took most annotators about 2 to 2.5 hours to rate 1,000 words. Table \ref{table:lex} shows few examples from the lexicon along with their human ratings.

In order to assess the annotation quality, we computed the Pearson correlation between each annotator's annotations and the average of the rest of the annotations \cite{Eneko2014}. For our final word-complexity lexicon, we took an average of the human ratings for each word, discarding those (about 3\%) that had a difference $\geq$ 2 from the mean of the rest of the ratings. The overall inter-annotator agreement improved from 0.55 to 0.64 after discarding the outlying ratings. For the majority of the disagreements, the ratings of one annotator and the mean of the rest were fairly close: the difference is $\leq$ 0.5 for 47\% of the annotations; $\leq$ 1.0 for 78\% of the annotations; and $\leq$ 1.5 for 93\% of the annotations on the 6-point scale. We hired annotators of different native languages intentionally, which may have contributed to the variance in the judgments.\footnote{One recent work similarly observed lower inter-annotator agreement among non-native speakers than native speakers when asked to identify complex words in given text paragraphs \cite{Muhie2017}.} We leave further investigation and possible crowdsourcing annotation to future work.
% Another recent work on sentiment analysis also suggested that annotator disagreement could be useful indicator of ambiguous or difficult cases rather than poor annotation quality \cite{kenyondean-EtAl:2018:N18-1}

%==========================
\section{Neural Readability Ranking Model for Words and Phrases}
\label{sec:model}

In order to predict the complexity of any given word or phrase, within or outside the lexicon, we propose a Neural Readability Ranking model that can leverage the created word-complexity lexicon and take context (if available) into account to further improve performance. Our model uses a Gaussian-based vectorization layer to exploit numerical features more effectively and can outperform the state-of-the-art approaches on multiple lexical simplification tasks with or without the word-complexity lexicon. We describe the general model framework in this section, and task-specific configurations in the experiment section (\S \ref{sec:experiments}).

\begin{figure*}
\begin{center}
\includegraphics[width=5.2in]{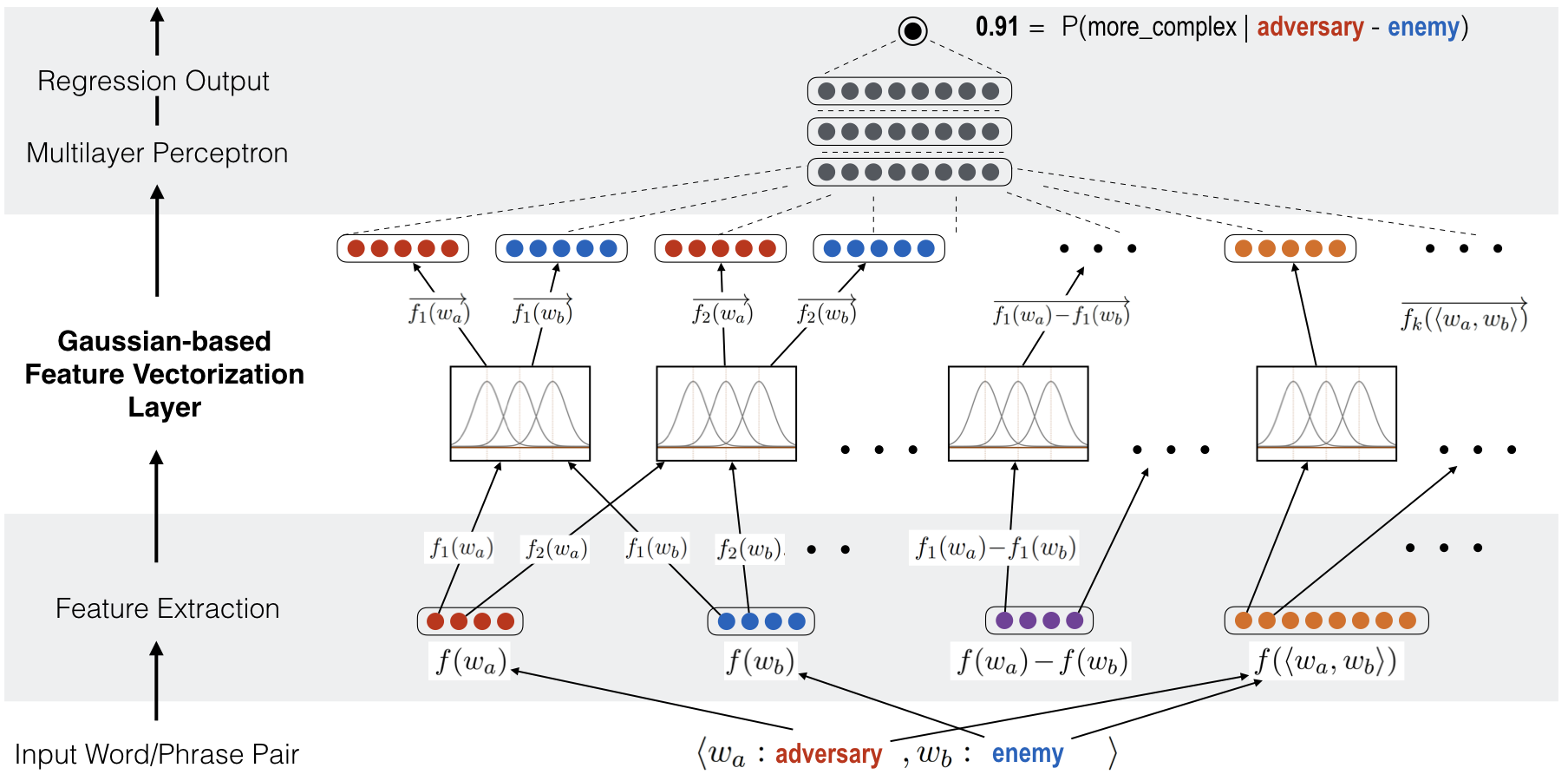}
\end{center}
\caption{The neural readability ranking (NRR) model.}
\label{fig:nn}
\end{figure*}

\subsection{Neural Readability Ranker (NRR)}

Given a pair of words/phrases $\langle w_a, w_b\rangle$ as input, our model aims to output a real number that indicates the relative complexity $P(y|\langle w_a, w_b\rangle)$ of $w_a$ and $w_b$. If the output value is negative, then $w_a$ is simpler than $w_b$ and vice versa. Figure \ref{fig:nn} shows the general architecture of our ranking model highlighting the three main components:

\begin{enumerate}
  \item An input \textbf{feature extraction layer} (\S \ref{sec:features}) that creates lexical and corpus-based features for each input $f(w_a)$ and $f(w_b)$, and pairwise features $f(\langle w_a,w_b \rangle)$. We also inject the word-complexity lexicon into the model as a numerical feature plus a binary indicator.

  \item A \textbf{Gaussian-based feature vectorization layer} (\S \ref{sec:gaussian}) that converts each numerical feature, such as the lexicon scores and n-gram probabilities, into a vector representation by a series of Gaussian radial basis functions.

  \item A feedforward neural network performing regression with one \textbf{task-specific output node} that adapts the model to different lexical simplification tasks (\S \ref{sec:experiments}).
\end{enumerate}

Our model first processes each input word or phrase in parallel, producing vectorized features. All the features are then fed into a joint feedforward neural network.

\subsection{Features}
\label{sec:features}
We use a combination of rating scores from the word-complexity lexicon, lexical and corpus features \cite{pavlick2016a} and collocational features \cite{paetzold2017a}. %We choose these features because they are simple, effective, and facilitate direct comparison of our method with prior work, namely the SimplePPDB \cite{pavlick2016a} and the neural ranker in \cite{paetzold2017a}.

%\footnote{For different experiments, we may use different subsets of applicable features, specified in details in \S \ref{}.}

We inject the word-complexity lexicon into the NRR model by adding two features for each input word or phrase: a 0-1 binary feature representing the presence of a word (the longest word in a multi-word phrase) in the lexicon, and the corresponding word complexity score. For out-of-vocabulary words, both features have the value 0. We back-off to the complexity score of the lemmatized word if applicable. We also extract the following features: phrase length in terms of words and characters, number of syllables, frequency with respect to Google Ngram corpus \cite{brants2006a}, the relative frequency in Simple Wikipedia with respect to normal Wikipedia \cite{pavlick2015a} and ngram probabilities from a 5-gram language model trained on the SubIMDB corpus \cite{paetzold2016c}, which has been shown to work well for lexical simplification. For a word $w$, we take language model probabilities of all the possible n-grams within the context window of 2 to the left and right of $w$. When $w$ is a multi-word phrase, we break $w$ into possible n-grams and average the probabilities for a specific context window.

For an input pair of words/phrases $\langle w_a, w_b\rangle$, we include individual features $f(w_1)$, $f(w_2)$ and the differences $f(w_a)\! -\!f(w_b)$. We also use pairwise features $f(\langle w_a, w_b\rangle)$ including cosine similarity $cos(\overrightarrow{w}_a, \overrightarrow{w}_b)$ and the difference $\overrightarrow{w}_a\! -\!\overrightarrow{w}_b$ between the word2vec \cite{mikolov2013} embedding of the input words. The embeddings for a mutli-word phrase are obtained by averaging the embeddings of all the words in the phrase. We use the 300-dimensional embeddings  pretrained on the Google News corpus, which is released as part of the word2vec package.\footnote{\url{https://code.google.com/archive/p/word2vec/}}

\subsection{Vectorizing Numerical Features via Gaussian Binning}
\label{sec:gaussian}

Our model relies primarily on numerical features as many previous approaches for lexical simplification. Although these continuous features can be directly fed into the network, it is helpful to exploit fully the nuanced relatedness between different intervals of feature values.

We adopt a smooth binning approach and project each numerical feature into a vector representation by applying multiple Gaussian radial basis functions. For each feature $f$, we divide its value range [$f_{min}$, $f_{max}$] evenly into $k$ bins and place a Gaussian function for each bin with the mean $\mu_{j}$ ($j \in \{1, 2, \dots, k\}$) at the center of the bin and standard deviation $\sigma$. We specify $\sigma$ as a fraction $\gamma$ of bin width:
\begin{equation}
\sigma = \frac{1}{k} (f_{max} - f_{min}) \cdot \gamma
\end{equation}
where $\gamma$ is a tunable hyperparameter in the model. For a given feature value $f(\cdot)$, we then compute the distance to each bin as follows:
\begin{equation}
d_{j}(f(\cdot)) = e^{-\frac{(f(\cdot)-\mu_{j})^2}{2 \sigma^2}}
\end{equation}
and normalize to project into a $k$-dimensional vector $\overrightarrow{f(\cdot)} = (d_1, d_2, \dots, d_k)$.

\begin{figure}[ht!]
\begin{center}
\includegraphics[width=2.5in]{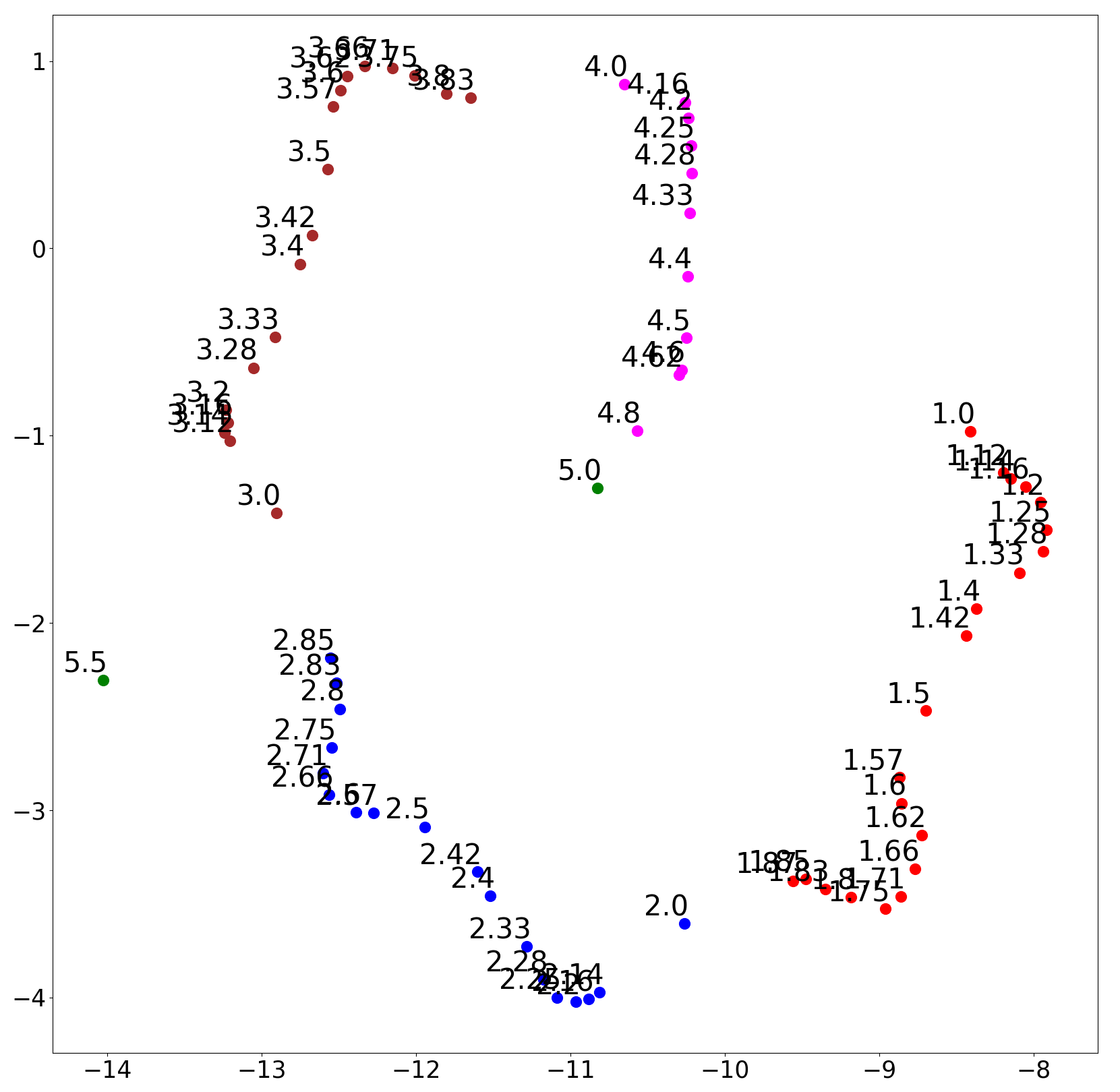}
\end{center}

\caption{t-SNE visualization of the complexity scores, ranging between 1.0 and 5.0, of 300 random words from the word-complexity lexicon vectorized into 10-dimensional representations by applying Gaussian radial basis functions.}
\label{fig:gb}
\end{figure}

We vectorize all the features except word2vec vectors,  $\overrightarrow{f(w_a)}$, $\overrightarrow{f(w_b)}$, $\overrightarrow{f(w_a)\! -\!f(w_b)}$, and $\overrightarrow{f(\langle w_a, w_b\rangle)}$, then concatenate them as inputs. Figure \ref{fig:gb} presents a motivating t-SNE visualization of the word-complexity scores from the lexicon after the vectorization in our NRR model, where different feature value ranges are gathered together with some distances in between.

%\begin{figure}[ht!]
%\begin{center}
%\includegraphics[width=3.0in, height=1.2in]{gaussian_binning.png}
%\end{center}
%\label{fig:gb_imp}
%\caption{Gaussian binning and feature vectorizaiton process for $k$ = 5 bins.}
%\end{figure}

\subsection{Training and Implementation Details}
We use PyTorch framework to implement the NRR model, which consists of an input layer, three hidden layers with eight nodes in each layer and the \textit{tanh} activation function, and a single node linear output layer. The training objective is to minimize the Mean Squared Error (MSE):
\begin{equation}
L(\theta) = \frac{1}{m} \sum_{i=1}^{m} (y_i - \hat{y}_i)^2
\end{equation}
where $y_i$ and $\hat{y}_i$ are the true and predicted relative complexity scores of $\langle w_a, w_b\rangle$ which can be configured accordingly for different lexical simplification tasks and datasets, $m$ is the number of training examples, and $\theta$ is the set of parameters of the NRR model. We use Adam algorithm \cite{Kingma2014} for optimization and also apply a dropout of 0.2 to prevent overfitting. We set the rate to 0.0005 and 0.001 for experiments in (\S \ref{sec:SR}) and (\S \ref{sec:ppdb}) respectively. For Gaussian binning layer, we set the number of bins $k$ to 10 and $\gamma$ to 0.2 without extensive parameter tuning. For each experiment,we report results with 100 epochs.

%==========================

\section{Lexical Simplification Applications}
\label{sec:experiments}

As the lexical simplification research field traditionally studies multiple sub-tasks and datasets, we present a series of experiments to demonstrate the effectiveness of our newly created lexicon and neural readability ranking (NRR) model.

\subsection{Substitution Ranking}
\label{sec:SR}

Given an instance consisting of a target complex word in a sentence and a set of candidate substitutions, the goal of the Substitution Ranking task is to rank the candidates in the order of their simplicity. In this section, we show that our proposed NRR model outperforms the state-of-the-art neural model on this task, with or without using the word-complexity lexicon.

\paragraph{Data.} We use the dataset from the English Lexical Simplification shared-task at SemEval 2012 \cite{specia2012} for evaluation. The training and test sets consist of 300 and 1,710 instances, respectively,  with a total of 201 target words (all single word, mostly polysemous) and each in 10 different sentences. One example of such instance contains a target complex word in context:

\vspace{.1in}
\textit{When you think about it, that's pretty \underline{terrible}}.
\vspace{.1in}

\noindent and a set of candidate substitutions \{\textit{bad}, \textit{awful}, \textit{deplorable}\}. Each instance contains at least 2 and an average of 5 candidates to be ranked. There are a total of 10034 candidates in the dataset, 88.5\% of which are covered by our word-complexity lexicon and 9.9\% are multi-word phrases (3438 unique candidates with 81.8\% in-vocabulary and 20.2\% multi-word).

\paragraph{Task-specific setup of the NRR model.}

We train the NRR model with every pair of candidates $\langle c_a,c_b \rangle$ in a candidate set as the input, and the difference of their ranks $r_a\! -\!r_b$ as the ground-truth label. For each such pair, we also include another training instance with $\langle c_b,c_a \rangle$ as the input and $r_b\! -\!r_a$ as the label. Given a test instance with candidate set $C$, we rank the candidates as follows: for every pair of candidates $\langle c_a,c_b \rangle$, the model predicts the relative complexity score $S(c_a, c_b)$; we then compute a single score $R(c_a) = \sum_{c_a\neq c_b \in C} S(c_a, c_b)$ for each candidate by aggregating pairwise scores and rank the candidates in the increasing order of these scores.

\paragraph{Comparison to existing methods.}  We compare with the state-of-the-art neural model \cite{paetzold2017a} for substitution ranking with the best reported results on the SemEval 2012 dataset. Our baselines also include several other existing methods: Biran et al. \shortcite{P11-2087}, Kajiwara et al. \shortcite{Kajiwara2013}, and Glava\v{s} \& \v{S}tajner \shortcite{glavavs-vstajner:2015:ACL-IJCNLP}, which use carefully designed heuristic scoring functions to combine various information such as corpus statistics and semantic similarity measures from WordNet; Horn et al. \shortcite{horn-manduca-kauchak:2014:P14-2} and the Boundary Ranker \cite{paetzold2015}, which respectively use a supervised SVM ranking model and pairwise linear classification model with various features. All of these methods have been implemented as part of the
LEXenstein toolkit \cite{paetzold2015}, which we use for the
experimental comparisons here. In addition, we also compare to the best system \cite{jauhar-specia:2012:STARSEM-SEMEVAL} among participants at SemEval 2012, which used SVM-based ranking.

%\footnote{\url{https://github.com/ghpaetzold/LEXenstein}}

\begin{table}[t!]
\centering
\small
\begin{tabular}{l|c|c}
\hline
& \textbf{P@1} & \textbf{Pearson} \\
\hline
Biran et al. \shortcite{P11-2087} & 51.3 & 0.505 \\
%Devlin & 0.596 & 0.614\\
Jauhar \& Specia \shortcite{jauhar-specia:2012:STARSEM-SEMEVAL} & 60.2 & 0.575 \\
Kajiwara et al. \shortcite{Kajiwara2013} & 60.4 & 0.649 \\
Horn et al. \shortcite{horn-manduca-kauchak:2014:P14-2} & 63.9 & 0.673 \\
Glava\v{s} \& \v{S}tajner \shortcite{glavavs-vstajner:2015:ACL-IJCNLP} & 63.2 & 0.644 \\
Boundary Ranker & 65.3 & 0.677  \\
Paetzold \& Specia \shortcite{paetzold2017a} & 65.6 & 0.679 \\
\hline
\hline
NRR$_{all}$ & 65.4 & 0.682 \\
NRR$_{all + binning}$ & 66.6 & 0.702* \\
NRR$_{all + binning + WC}$ & \textbf{67.3}* & \textbf{0.714}* \\
\hline
\end{tabular}
\small
\caption{Substitution Ranking evaluation on English Lexical Simplification shared-task of SemEval 2012. P@1 and Pearson correlation of our neural readability ranking (NRR) model compared to the state-of-the-art neural model \cite{paetzold2017a} and other methods.  $\ast$ indicates statistical significance ($p < 0.05$) compared to the best performing baseline \cite{paetzold2017a}.}
\label{table:sr}
\end{table}

\paragraph{Results.}
Table \ref{table:sr} compares the performances of our NRR model to the state-of-the-art results reported by Paetzold and Specia \shortcite{paetzold2017a}. We use precision of the simplest candidate (P@1) and Pearson correlation to measure performance. P@1 is equivalent to TRank \cite{specia2012}, the official metric for the SemEval 2012 English Lexical Simplification task. While P@1 captures the practical utility of an approach, Pearson correlation indicates how well the system's rankings correlate with human judgment. We train our NRR model with all the features (NRR$_{all}$) mentioned in \S \ref{sec:features} except the word2vec embedding features to avoid overfitting on the small training set. Our full model (NRR$_{all + binning + WC}$) exhibits a statistically significant improvement over the state-of-the-art for both measures. We use paired bootstrap test \cite{KirkpatrickBK12, EfroTibs93} as it can be applied to any performance metric. We also conducted ablation experiments to show the effectiveness of the Gaussian-based feature vectorization layer ($_{+binning}$) and the word-complexity lexicon ($_{+WC}$).

\begin{table}[t!]
\begin{small}

\begin{tabular}{p{1.8cm} p{5.1cm}}
\multicolumn{2}{l}{paraphrases of \textit{\textbf{`modification'}} ranked by simplicity} \\
\hline
SimplePPDB & \textit{tweak, modify, process, variable, layout}\\
SimplePPDB++ & \textit{change, adjustment, amendment, shift, difference} \\
\end{tabular}\\
\vspace{1mm}

\begin{tabular}{p{1.8cm} p{5.5cm}}
\multicolumn{2}{l}{paraphrases of \textit{\textbf{`aggregation'}}} \\
\hline
SimplePPDB & \textit{pod,	swarm, node, clump, pool}\\
SimplePPDB++ & \textit{cluster, pool, collection, addition, grouping} \\
\end{tabular}\\
\vspace{1mm}

\begin{tabular}{p{1.8cm} p{5.1cm}}
\multicolumn{2}{l}{paraphrases of \textit{\textbf{`of transnational corporation'}}} \\
\hline
SimplePPDB & \textit{of corporation, by corporation, of enterprise, of tncs, of business}\\
SimplePPDB++ & \textit{of business, of firm, of corporation, of company, of enterprise} \\
\end{tabular}
\vspace{1mm}

\begin{tabular}{p{1.8cm} p{5.1cm}}
\multicolumn{2}{l}{paraphrases of \textit{\textbf{`should reject'}}} \\
\hline
SimplePPDB & \textit{refuse, discard, repudiate, shun, dismiss}\\
SimplePPDB++ & \textit{vote against, set aside, throw out, say no to, turn away} \\
\end{tabular}
\vspace{1mm}

\caption{SimplePPDB++ includes lexical and phrasal paraphrases with improved readability ranking scores by our NRR$_{all + binning + WC}$ model. Shown are the top 5 ranked simplifications according to SimplePPDB++ for several input words/phrases, in comparison to the previous work of SimplePPDB \cite{pavlick2016a}.}
\label{table:sppdbeg}
\end{small}
\end{table}

\subsection{SimplePPDB++}
\label{sec:ppdb}

We also can apply our NRR model to rank the lexical and phrasal paraphrase rules in the Paraphrase Database (PPDB) \cite{pavlick2015b}, and identify good simplifications (see examples in Table \ref{table:sppdbeg}). The resulting lexical resource, SimplePPDB++, contains all 13.1 million lexical and phrasal paraphrase rules in the XL version of PPDB 2.0 with readability scores in `simplifying', `complicating', or `nonsense/no-difference' categories, allowing flexible trade-off between high-quality and high-coverage paraphrases. In this section, we show the effectiveness of the NRR model we used to create SimplePPDB++ by comparing with the previous version of SimplePPDB \cite{pavlick2016a} which used a three-way logistic regression classifier. In next section, we demonstrate the utility of SimplePPDB++ for the Substitution Generation task.

\paragraph{Task-specific setup of NRR model.}
 We use the same manually labeled data of 11,829 paraphrase rules as SimplePPDB for training and testing, of which 26.5\% labeled as `simplifying', 26.5\% as `complicating', and 47\% as `nonsense/no-difference'. We adapt our NRR model to perform the three-way classification by treating it as a regression problem. During training, we specify the ground truth label as follows: $y$ = -1 if the paraphrase rule belongs to the `complicating' class, $y$ = +1 if the rule belongs to the `simplifying'class, and $y$ = 0 otherwise. For predicting, the network produces a single real-value output $\hat{y}\! \in\! [-1, 1] $ which is then mapped to three-class labels based on the value ranges for evaluation. The thresholds for the value ranges are -0.4 and 0.4 chosen by cross-validation.

\begin{table}[t!]
\centering
\small
\begin{tabular}{l|l|l|l}
\hline
& \textbf{Acc.} & \textbf{P$_{+1}$} & \textbf{P$_{-1}$} \\
\hline
Google Ngram Frequency & 49.4 & 53.7 & 54.0 \\
Number of Syllables & 50.1 & 53.8 & 53.3 \\
Character \& Word Length & 56.2 & 55.7 & 56.1 \\
W2V & 60.4 & 54.9 & 53.1 \\
SimplePPDB &  62.1 & 57.6 & 57.8 \\
\hline
\hline
NRR$_{all}$ & 59.4 & 61.8 & 57.7\\
NRR$_{all + binning}$ & 64.1 & 62.1 & 59.8 \\
NRR$_{all + binning + WC}$ &  \textbf{65.3}* & \textbf{65.0}* & \textbf{61.8}* \\
\hline
\end{tabular}
\small
\caption{Cross-validation accuracy and precision of our neural readability ranking (NRR) model used to create SimplePPDB++, in comparison to the SimplePPDB and other baselines. P$_{+1}$ stands for the precision of `simplifying' paraphrase rules and P$_{-1}$ for the precision of `complicating' rules. * indicates statistical significance ($p < 0.05$) compared to the best performing baseline \cite{pavlick2016a}.}
\label{table:sppdb}
\end{table}

\paragraph{Comparison to existing methods.}

We compare our neural readability ranking (NRR) model used to create the SimplePPDB++ against SimplePPDB, which uses a multi-class logistic regression model. We also use several other baselines, including W2V which uses logistic regression with only word2vec embedding features.

\paragraph{Results.}
Following the evaluation setup in previous work \cite{pavlick2016a}, we compare accuracy and precision by 10-fold cross-validation. Folds are constructed in such a way that the training and test vocabularies are disjoint. Table \ref{table:sppdb} shows the performance of our model compared to SimplePPDB and other baselines. We use all the features (NRR$_{all}$) in \S \ref{sec:features} except for the context features as we are classifying paraphrase rules in PPDB that come with no context. SimplePPDB used the same features plus additional discrete features, such as POS tags, character unigrams and bigrams. Our neural readability ranking model alone with Gaussian binning (NRR$_{all + binning}$) achieves better accuracy and precision while using less features. Leveraging the lexicon (NRR$_{all + binning + WC}$) shows statistically significant improvements over SimplePPDB rankings based on the paired bootstrap test. The accuracy increases by 3.2 points, the precision for `simplifying'  class improves by 7.4 points and the precision for `complicating' class improves by 4.0 points.
%Gaussian binning and feature vectorization is shown to be helpful for exploiting the numerical features more effectively.

\subsection{Substitution Generation}

Substitution Generation is arguably the most challenging research problem in lexical simplification, which involves producing candidate substitutions for each target complex word/phrase, followed by the substitution ranking. The key focus is to not only have better rankings, but more importantly, to have a larger number of simplifying substitutions generated. This is a more realistic evaluation to demonstrate the utility of SimplePPDB++ and the effectiveness of the NRR ranking model we used to create it, and how likely such lexical resources can benefit developing end-to-end sentence simplification system \cite{narayan-gardent:2016:INLG,zhang-lapata:2017:EMNLP2017} in future work.

\paragraph{Data.} We use the dataset from \cite{pavlick2016a}, which contains 100 unique target words/phrases sampled from the Newsela Simplification Corpus \cite{Xu-EtAl:2015:TACL} of news articles, and follow the same evaluation procedure. We ask two annotators to evaluate whether the generated substitutions are good simplifications.

\paragraph{Comparison to existing methods.}  We evaluate the correctness of the substitutions generated by SimplePPDB++ in comparison to several existing methods: Glava\v{s} \cite{glavavs-vstajner:2015:ACL-IJCNLP}, Kauchak \cite{Coster:2011:LSS:2107679.2107680}, WordNet Generator \cite{Devlin1998,E99-1042}, and SimplePPDB \cite{pavlick2016a}. Glava\v{s} obtains candidates with the highest similarity scores in the GloVe \cite{pennington2014glove} word vector space. Kauchak's generator is based on Simple Wikipedia and normal Wikipedia parallel corpus and automatic word alignment. WordNet-based generator simply uses the synonyms of word in WordNet \cite{Miller:1995:WLD:219717.219748}. For all the existing methods, we report the results based on the implementations in \cite{pavlick2016a}, which used SVM-based ranking. For both SimplePPDB and SimplePPDB++, extracted candidates are high quality paraphrase rules (quality score $\geq$3.5 for words and $\geq$4.0 for phrases) belonging to the same syntactic category as target word according to PPDB 2.0 \cite{pavlick2015b}.

\paragraph{Results.} Table \ref{table:sg} shows the comparison of SimplePPDB and SimplePPDB++ on the number of substitutions generated for each target, the mean average precision and precision@1 for the final ranked list of candidate substitutions. This is a fair and direct comparison between SimplePPDB++ and SimplePPDB, as both methods have access to the same paraphrase rules in PPDB as potential candidates. The better NRR model we used in creating SimplePPDB++ allows improved selections and rankings of simplifying paraphrase rules than the previous version of SimplePPDB. As an additional reference, we also include the measurements for the other existing methods based on \cite{pavlick2016a}, which, by evaluation design, are focused on the comparison of precision while PPDB has full coverage.

%As an additional reference, we also include the measurements for the other existing methods based on \cite{pavlick2016a}, which, by evaluation design, are focused on the unbiased comparison on the precision while PPDB has full coverage of the list of 100 targets.

\begin{table}[t!]
\centering
\small
\begin{tabular}{p{3.4cm}|c|c|c}
\hline
& \textbf{\#PPs} & \textbf{MAP} & \textbf{P@1} \\
\hline
Glava\v{s}$_{(n=95)}$ & --- & 22.8 & 13.5 \\
WordNet$_{(n=82)}$ & 6.63 & 62.2 & 50.6\\
Kauchak$_{(n=48)}$ & 4.39 & \textbf{76.4}$^\dagger$ & 68.9 \\
SimplePPDB$_{(n=100)}$ & 8.77 & 67.8 & 78.0 \\
\hline
\hline
SimplePPDB++$_{(n=100)}$ & \textbf{9.52} & 69.1 & \textbf{80.2}  \\
\hline

\end{tabular}
\small
\caption{Substitution Generation evaluation with Mean Average Precision, Precision@1 and the average number of paraphrases generated per target for each method. $n$ is the number of target complex words/phrases for which the model generated $>0$ candidates. Kauchak$^\dagger$ has an advantage on MAP because it generates the least number of candidates. Glava\v{s} is marked as `-' because it can technically generate as many words/phrases as are in the vocabulary.}
\label{table:sg}
\end{table}

\begin{table*}
\centering
\small
\begin{tabular}{l|c|c|c||c|c|c}
\hline
& \multicolumn{3}{c||}{\textbf{CWI SemEval 2016}} & \multicolumn{3}{c}{\textbf{CWIG3G2 2018}}
\\ \cline{2-4}  \cline{5-7}
& \textbf{G-score} & \textbf{F-score} &\textbf{Accuracy}
& \textbf{G-score} & \textbf{F-score} &\textbf{Accuracy}
\\ \hline
Length & 47.8 & 10.7 & 33.2 & 70.8 & 65.9 & 67.7 \\
Senses & 57.9 & 12.5 & 43.6 & 67.7 & 62.3 & 54.1 \\
SimpleWiki & 69.7 & 16.2 & 58.3 & 73.1 & 66.3 & 61.6 \\
NearestCentroid  & 66.1 & 14.8 & 53.6 & 75.1 & 66.6 & 76.7 \\
SV000gg  & \underline{77.3} & 24.3 & 77.6 & 74.9 & \underline{73.8} & \underline{78.7} \\
\hline
\hline
$WC$-only & 68.5 & \textbf{30.5} & \textbf{87.7} & 71.1 & 67.5 & 69.8 \\
NearestCentroid$_{+WC}$ & 70.2 & 16.6 & 61.8 & \textbf{77.3} & 68.8 & 78.1\\
SV000gg$_{+WC}$ & \textbf{78.1} & \underline{26.3} & \underline{80.0} & \underline{75.4} & \textbf{74.8} & \textbf{80.2} \\
\hline
\end{tabular}
\small
\caption{Evaluation on two datasets for English complex word identification. Our approaches that utilize the word-complexity lexicon (${WC}$) improve upon the nearest centroid \cite{Muhie2017} and SV000gg \cite{paetzold2016b} systems. The best performance figure of each column is denoted in \textbf{bold} typeface and the second best is denoted by an \underline{underline}.}
\label{table:cwi}
\end{table*}

\begin{table}[h!]
\centering
\small
\begin{tabular}{c|c|c}
\hline
\textbf{CWI SemEval 2016} & total (IV\%) & unique (IV\%)  \\
\hline
simple & 85621 (94.7\%) & 14129 (77.6\%)  \\
complex & 4837 (57.4\%) & 3836 (54.8\%)  \\
\hline
\hline
\textbf{CWIG3G2} & total (IV\%) & unique (IV\%)  \\
\hline
simple & 20451 (89.8\%) &  5576 (82.1\%) \\
complex & 14428 (81.1\%) &  8376 (76.0\%) \\
\hline
\end{tabular}
\small
\caption{Statistics of CWI datasets -- total number of target words/phrases, number of unique targets, and in-vocabulary (IV) ratio with respect to our word-complexity lexicon.}
\label{table:cwi_data}
\end{table}

\subsection{Complex Word Identification}
Complex Word Identification (CWI) identifies the difficult words in a sentence that need to be simplified. According to Shardlow \shortcite{shardlow2014}, this step can improve the simplification system by avoiding mistakes such as overlooking challenging words or oversimplifying simple words. In this section, we demonstrate how our word-complexity lexicon helps with the CWI task by injecting human ratings into the state-of-the-art systems.

\paragraph{Data.} The task is to predict whether a target word/phrase in a sentence is `simple' or `complex', and an example instance is as follows:

\vspace{.1in}
\textit{ Nine people were killed in the \underline{bombardment}.}
\vspace{.1in}

\noindent We conduct experiments on two datasets: (i) Semeval 2016 CWI shared-task dataset \cite{paetzold2016a}, which has been widely used for evaluating CWI systems and contains 2,237 training and 88,221 test instances from Wikipedia; and (ii) CWIG3G2 dataset \cite{Muhie2017}, which is also known as English monolingual CWI 2018 shared-task dataset \cite{Muhie2018} and comprises of 27,299 training, 3,328 development and 4,252 test instances from Wikipedia and news articles. Table \ref{table:cwi_data} shows the coverage of our word-complexity lexicon over the two CWI datasets.

\paragraph{Comparison to existing methods.}

We consider two state-of-the-art CWI systems: (i) the nearest centroid classifier proposed in \cite{Muhie2017}, which uses phrase length, number of senses, POS tags, word2vec cosine similarities, n-gram frequency in Simple Wikipedia corpus and Google 1T corpus as features; and (ii) SV000gg \cite{paetzold2016b} which is an ensemble of binary classifiers trained with a combination of lexical, morphological, collocational, and semantic features. The latter is the best performing system on the Semeval 2016 CWI dataset. We also compare to threshold-based baselines that use word length, number of word senses and frequency in the Simple Wikipedia.

\paragraph{Utilizing the word-complexity lexicon.}

We enhance the SV000gg and the nearest centroid classifier by incorporating the word-complexity lexicon as additional features as described in \S \ref{sec:features}. We added our modifications to the implementation of SV000gg in the LEXenstein toolkit, and used our own implementation for the nearest centroid classifier. Additionally, to evaluate the word-complexity lexicon in isolation, we train a decision tree classifier with only human ratings as input ($WC$-only), which is equivalent to learning a threshold over the human ratings.

\paragraph{Results.}

We compare our enhanced approaches (SV000gg$_{+WC}$ and NC$_{+WC}$) and lexicon only approach ($WC$-only), with the state-of-the-art and baseline threshold-based methods. For measuring performance, we use F-score and accuracy as well as G-score, the harmonic mean of accuracy and recall. G-score is the official metric of the CWI task of Semeval 2016. Table \ref{table:cwi} shows that the word-complexity lexicon improves the performance of SV000gg and the nearest centroid classifier in all the three metrics. The improvements are statistically significant according to the paired bootstrap test with $p < 0.01$. The word-complexity lexicon alone ($WC$-only) performs satisfactorily on the CWIG3G2 dataset, which effectively is a simple table look-up approach with extreme time and space efficiency. For CWI SemEval 2016 dataset, $WC$-only approach gives the best accuracy and F-score, though this can be attributed to the skewed distribution of dataset (only 5\% of the test instances are `complex').

\section{Related Work}
\label{sec:relatedwork}

\paragraph{Lexical simplification:}
Prior work on lexical simplification depends on lexical and corpus-based features to assess word complexity. For complex word identification, there are broadly two lines of research: learning a frequency-based threshold over a large corpus \cite{Shardlow2013a} or training an ensemble of classifiers over a combination of lexical and language model features \cite{Shardlow2013b, paetzold2016a, Muhie2017, Kriz2018:NAACL}. Substitution ranking also follows similar trend.
Biran et al. \shortcite{P11-2087} and Bott et al. \shortcite{Bott2012} employed simplicity measures based on word length and word frequencies from Wikipedia and Simple Wikipedia. Kajiwara et al. \shortcite{Kajiwara2013} combined WordNet similarity measures with Simple Wikipedia frequencies. Glava\v{s} and \v{S}tajner \shortcite{glavavs-vstajner:2015:ACL-IJCNLP} averaged the rankings produced by a collection of frequency, language model and semantic similarity features. Horn et al. \shortcite{horn-manduca-kauchak:2014:P14-2} trained an SVM classifier over corpus-based features.

Only recently, researchers started to apply neural networks to simplification tasks. To the best of our knowledge, the work by Paetzold and Specia \shortcite{paetzold2017a} is the first neural model for lexical simplification which uses a feedforward network with language model probability features. Our NRR model is the first pairwise neural ranking model to vectorize numeric features and to embed human judgments using a word-complexity lexicon of 15,000 English words.

Besides lexical simplification, another line of relevant research is sentence simplification that uses statistical or neural machine translation (MT) approaches \cite{Xu-EtAl:2016:TACL,nisioi-EtAl:2017:Short,zhang-lapata:2017:EMNLP2017,vu-EtAl:2018:NAACL,guo-pasunuru-bansal:2018:C18-1}. It has shown possible to integrate paraphrase rules in PPDB into statistical MT for sentence simplification \cite{Xu-EtAl:2016:TACL} and bilingual translation \cite{mehdizadehseraj-siahbani-sarkar:2015:EMNLP}, while how to inject SimplePPDB++ into neural MT remains an open research question.

\paragraph{Lexica for simplification:} There have been previous attempts to use manually created lexica for simplification. For example, Elhadad and Sutaria \shortcite{Elhadad2007} used UMLS lexicon \cite{Bodenreider2007}, a repository of technical medical terms; Ehara et al. \shortcite{Ehara:2010:PRS:1719970.1719978} asked non-native speakers to answer multiple-choice questions corresponding to 12,000 English words to study each user's familiarity of vocabulary; Kaji et al. \shortcite{nobuhiro2012} and Kajiwara et al. \shortcite{Kajiwara2013} used a dictionary of 5,404 Japanese words based on the elementary school textbooks; Xu et al. \shortcite{Xu-EtAl:2016:TACL} used a list of 3,000 most common English words; Lee and Yeung \shortcite{lee-yeung:2018:C18-1} used an ensemble of vocabulary lists of different complexity levels. However, to the best of our knowledge, there is no previous study on manually building a large word-complexity lexicon with human judgments that has shown substantial improvements on automatic simplification systems. We were encouraged by the success of the word-emotion lexicon \cite{Mohammad13} and the word-happiness lexicon \cite{10.1371/journal.pone.0026752,dodds2015a}.

\paragraph{Vectorizing features:} Feature binning is a standard feature engineering and data processing method to discretize continuous values, more commonly used in non-neural machine learning models. Our work is largely inspired by recent works on entity linking that discussed feature quantization for neural models \cite{sil2017neural,DBLP:journals/corr/LiuLZWJ16} and neural dependency parsing with embeddings of POS tags as features \cite{chen-manning:2014:EMNLP2014}.

\section{Conclusion}
We proposed a new neural readability ranking model and showed significant performance improvement over the state-of-the-art on various lexical simplification tasks. We release a manually constructed word-complexity lexicon of 15,000 English words and an automatically constructed lexical resource, SimplePPDB++, of over 10 million paraphrase rules with quality and simplicity ratings. For future work, we would like to extend our lexicon to cover specific domains, different target users and languages.

\section*{Acknowledgments}
We thank anonymous reviewers for their thoughtful comments. We thank Avirup Sil and Anastasios Sidiropoulos for valuable discussions, Sanja \v{S}tajner and Seid Muhie Yimam for sharing their code and data. We also thank the annotators: Jeniya Tabassum, Ashutosh Baheti, Wuwei Lan, Fan Bai, Alexander Konovalov, Chaitanya Kulkarni, Shuaichen Chang, Jayavardhan Reddy, Abhishek Kumar and Shreejit Gangadharan.

This material is based on research sponsored by
the NSF under grants IIS-1822754 and IIS-1755898. The views and conclusions contained in this publication are those
of the authors and should not be interpreted as representing official policies or endorsements of the NSF or the U.S. Government.

\bibliography{emnlp2018}
\bibliographystyle{emnlp2018}
\end{document}